\documentclass[11pt,twocolumn]{article}
\usepackage[utf8]{inputenc}
\usepackage{times}
\usepackage{graphicx}
\usepackage{amsmath, amssymb}
\usepackage{hyperref}
\usepackage{geometry}
\usepackage[ruled,vlined]{algorithm2e}
\usepackage[numbers,sort&compress]{natbib}
\usepackage{fontawesome5}
\usepackage[table]{xcolor}
\usepackage{booktabs}
\usepackage{enumitem}
\usepackage{placeins}
\usepackage{stfloats}
\usepackage{capt-of}
\setlist{nosep}

\title{SatEdit: Mask-Conditioned Image Editing via VLM-Guided Segment Annotation}

\author{
    Muhammad Talha \\
    Independent Researcher \\
    \texttt{trizwan24@gmail.com}
    \and
    Muhammad Ahmed Amer \\
    Independent Researcher \\
    \texttt{iammaa2001@gmail.com}
}

\date{}

\begin{document}

\twocolumn[{
\maketitle
\vspace{-1.5em}

\begin{center}
\large
\textbf{
\faGlobe\ \href{https://muhammad-talha-ad.github.io/SatEdit/}{Project Page}
\hspace{2em}
\faGithub\ \href{https://github.com/muhammad-talha-ad/SatEdit}{Code}
\hspace{2em}
\faDatabase\ \href{https://huggingface.co/MTalha2001/SatEdit}{Model}
}
\end{center}

\vspace{0.8em}
\begin{center}
\makebox[\textwidth][c]{\includegraphics[width=1.10\textwidth]{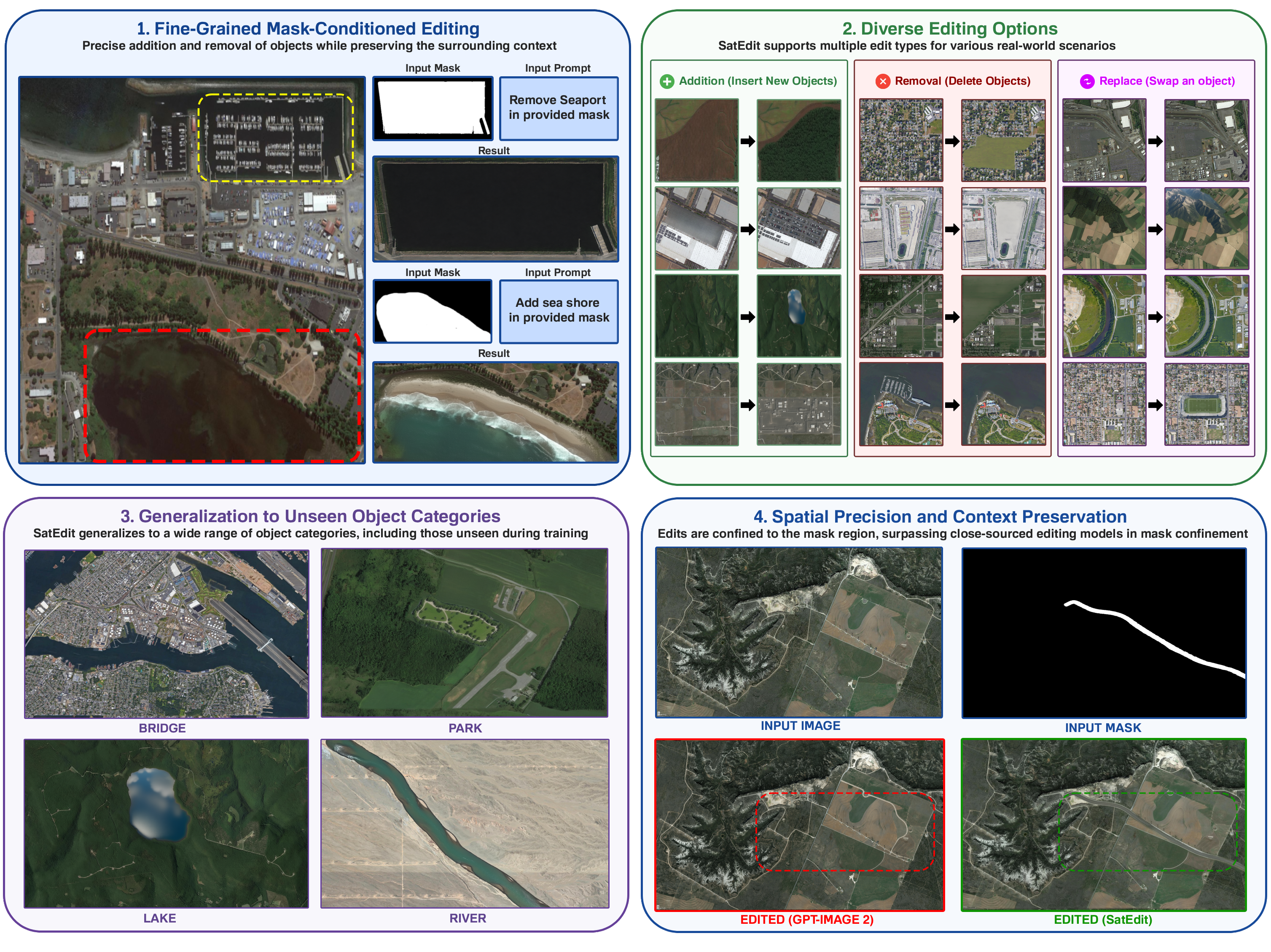}}
\refstepcounter{figure}\label{fig:overview}
\parbox{0.90\textwidth}{\small \centering Figure~\thefigure: \textbf{Overview of SatEdit.} Given a satellite image, binary mask, and text prompt, SatEdit edits the selected region while preserving the surrounding scene. The model is trained on object addition and removal, and we also observe qualitative transfer to replacement prompts and categories outside the training set.}
\end{center}

\vspace{1.0em}
}]

\begin{abstract}
    Satellite image editing requires spatially precise object-level control, but supervised editing datasets for overhead imagery are costly to build because object masks, semantic labels, and paired edits are rarely available at scale. We introduce \textbf{\emph{SatEdit}}, a mask-conditioned satellite image editing framework that constructs training supervision from unlabeled imagery. SatEdit proposes object masks with a segmentation foundation model, assigns semantic labels to sampled segments with a Vision-Language Model, and applies lightweight human verification before generating paired addition and removal examples through mask-guided inpainting. We fine-tune a high-resolution image editing backbone with LoRA on a SODA-A-derived dataset containing 1,014 images and 852 verified object annotations across 91 classes. In controlled comparisons with open-source and proprietary image editing models, SatEdit achieves the highest aggregate masked-region semantic alignment, with a CLIP score of 0.6322 and CLIP delta of 0.0726, while preserving the surrounding scene qualitatively. These results suggest that VLM-assisted segment annotation is a practical route to data-efficient, spatially controllable satellite image editing.
\end{abstract}

\section{Introduction}
\label{sec:intro}

Editing satellite imagery is useful only when the edit is both semantic and spatially local. Adding a vehicle, removing a building, or replacing a land-use object should change the target region while leaving roads, shadows, field boundaries, and nearby structures intact. This requirement is difficult for generic image editing models. Objects in overhead imagery are often small, visually compressed, and densely arranged, and the same semantic class can vary substantially with resolution, geography, and sensor conditions.

The main bottleneck is supervision. Most satellite datasets are designed for recognition: they provide image labels, boxes, or segmentation masks, but not paired before/after examples for instruction-guided editing. Natural-image editing corpora do contain such pairs, yet their object scales, viewpoints, and spatial priors differ from satellite scenes. As a result, models trained on general editing data can satisfy the text prompt while drifting outside the mask or producing objects whose scale, texture, or orientation is inconsistent with overhead imagery.

We propose \textbf{SatEdit}, a mask-conditioned framework for object-level satellite image editing. Rather than collecting editing pairs manually, SatEdit turns unlabeled satellite images into training supervision. The pipeline uses SAM2 \cite{ravi2024sam2} to generate candidate object masks, queries a Vision-Language Model on sampled masked regions to assign semantic labels, and uses a lightweight human verification pass to correct remaining label errors. The verified segments are then used to build paired object addition and removal examples through mask-guided inpainting.

The central idea is to replace expensive manual editing-pair annotation with segment-level supervision: masks provide spatial support, VLM labels provide object semantics, and inpainting converts verified segments into paired editing examples. This design decomposes supervision into spatial support, object semantics, label verification, and edited-image synthesis. We fine-tune a high-resolution image editing backbone with Low-Rank Adaptation (LoRA) on a SODA-A-derived dataset containing 1,014 images and 852 verified object annotations across 91 classes. At training time, the model receives the source image, text instruction, and explicit binary mask, so the desired semantic change is tied to a specified region.

We evaluate SatEdit against open-source and proprietary image editing systems on controlled mask-conditioned addition and removal tasks. SatEdit obtains the highest aggregate masked-region semantic alignment in our benchmark, with a CLIP score of 0.6322 and CLIP delta of 0.0726, while revealing a measurable tradeoff between edit strength, leakage, and background preservation. Qualitative results show the same pattern: the model places or removes objects inside the requested mask while preserving the surrounding scene, and also transfers to replacement prompts despite not being trained on replacement pairs.

We release the SatEdit pipeline and generated dataset to support reproducible work on data-efficient satellite image editing.

\clearpage
\textbf{Contributions.}
\begin{itemize}
\item We introduce SatEdit, a mask-conditioned satellite image editing framework for object-level addition and removal that preserves geographic context while localizing edits to user-specified regions.
\item We develop a scalable supervision pipeline that converts unlabeled overhead imagery into paired editing examples by combining SAM2 mask proposals \cite{ravi2024sam2}, VLM-based segment labeling, lightweight human verification, and mask-guided inpainting.
\item We construct a SODA-A-derived editing dataset with 1,014 images, 852 verified object annotations, and paired addition/removal examples spanning 91 semantic classes.
\item We evaluate SatEdit against open-source and proprietary editing systems, showing stronger masked-region semantic alignment and characterizing the tradeoff between edit strength, leakage, and background preservation.
\end{itemize}

\section{Related Work}
\label{sec:related}

\subsection{Vision-Language Modeling in Remote Sensing}

Vision-Language Models (VLMs) are increasingly used in remote sensing because they connect overhead visual patterns with open-vocabulary textual descriptions. CLIP \cite{radford2021clip} established contrastive language-image pretraining as a basis for zero-shot recognition, and subsequent geospatial work adapts this alignment to satellite imagery.

One line of work improves image-text representations for Earth observation. GRAFT \cite{graft2023} extends CLIP-based alignment by connecting overhead and ground-level views, supporting zero-shot classification, segmentation, retrieval, and visual question answering. SenCLIP \cite{senclip2023} improves zero-shot land-use and land-cover mapping with Sentinel-2 imagery and remote-sensing prompts. GeoVision Labeler \cite{geovision2023} similarly uses language-driven supervision to map semantic categories to satellite imagery.

A second line of work builds datasets and taxonomies for remote-sensing vision-language learning. Landsat30-AU \cite{landsat30au2023} introduces a large-scale vision-language dataset for Landsat imagery, and recent surveys \cite{survey2024} organize the area into contrastive, instruction-tuned, and generative paradigms.

These efforts mainly use VLMs as recognition, retrieval, mapping, or question-answering models. SatEdit uses VLMs as a supervision mechanism instead: segment-level VLM predictions become labels in a data-generation pipeline that trains a mask-conditioned image editor.

\subsection{Satellite Image Editing}

Satellite image editing is less developed than natural image editing because overhead imagery couples semantic changes with strict spatial and geographic constraints. The edited object must match the requested class, scale, orientation, and local context, while the surrounding scene should remain unchanged.

Early learning-based approaches often use Generative Adversarial Networks (GANs) for land-cover translation, seasonal transformation, and object-level manipulation \cite{isola2017pix2pix, zhu2017cyclegan, ganedit2022}. These methods show that generative models can alter remote-sensing imagery, but they usually provide limited object-level control and can struggle across scenes, resolutions, and sensor conditions.

Diffusion models improve the realism and controllability of image generation and editing. In remote sensing, ClimSat \cite{climsat2023} introduces a diffusion autoencoder for climate-conditioned satellite image editing, enabling transformations conditioned on environmental factors. RSEdit \cite{zhenyuan2026rsedit} studies text-guided image editing for remote sensing and compares conditioning strategies for adapting off-the-shelf text-to-image models to instruction-faithful edits that preserve geospatial structure. This line of work demonstrates the promise of generative satellite editing, but it still depends on curated supervision and does not directly address how to obtain large-scale object-level editing pairs from unlabeled imagery.

Recent work also studies spatially controlled satellite image synthesis. TerraDiT-$\Omega$ \cite{wei2026terradit} generates satellite imagery from native geospatial primitives such as polygons, polylines, bounding boxes, and points, using geometry-aware local attention to support controllable layouts across different annotation budgets. SatEdit is complementary: rather than synthesizing full satellite scenes from geospatial primitives, it edits an existing image under a binary mask and text instruction, and focuses on constructing the paired supervision needed for that setting.

Image enhancement and super-resolution methods, including EDSR \cite{lim2017edsr}, ESPCN \cite{shi2016espcn}, FSRCNN \cite{dong2016fsrcnn}, and LapSRN \cite{lai2017lapsrn}, solve a related but different problem. They improve reconstruction fidelity, whereas SatEdit targets semantic modification under an explicit mask and instruction. These methods therefore do not provide the masks, semantic labels, or paired before/after examples needed for object insertion and removal.

The key bottleneck for satellite image editing is scalable training data with precise spatial support and semantic labels. SatEdit addresses this bottleneck by constructing mask-conditioned editing supervision from unlabeled satellite imagery.

\subsection{Segmentation and Inpainting for Data Generation}

Segmentation and inpainting provide complementary tools for turning unlabeled images into editing supervision. The Segment Anything Model (SAM) \cite{kirillov2023sam} enables automatic decomposition of images into object-level regions without task-specific training, and SAM2 extends promptable segmentation with improved image and video support \cite{ravi2024sam2}. These properties make segmentation foundation models useful for proposing spatial supports in large satellite collections.

Inpainting models provide the second ingredient: they can synthesize plausible content inside a specified region. Methods such as LaMA \cite{suvorov2022lama} fill large missing regions with semantically consistent content and have been widely used for object removal and synthetic data generation.

However, segmentation alone is class-agnostic, VLM labeling alone does not produce edited images, and inpainting alone does not yield instruction-labeled training pairs. SatEdit combines these components into a single pipeline: segmentation proposes masks, VLMs assign object semantics, human verification corrects labels, and inpainting converts verified segments into paired addition and removal examples for satellite image editing.

\section{Method: SatEdit Framework}
\label{sec:method}

SatEdit turns unlabeled satellite images into mask-conditioned editing data, then uses that data to fine-tune an image editor. The method has three parts: segment proposal and labeling, construction of paired addition and removal examples, and LoRA fine-tuning of the editing model.

\begin{figure*}[!t]
\centering
\includegraphics[width=0.96\textwidth]{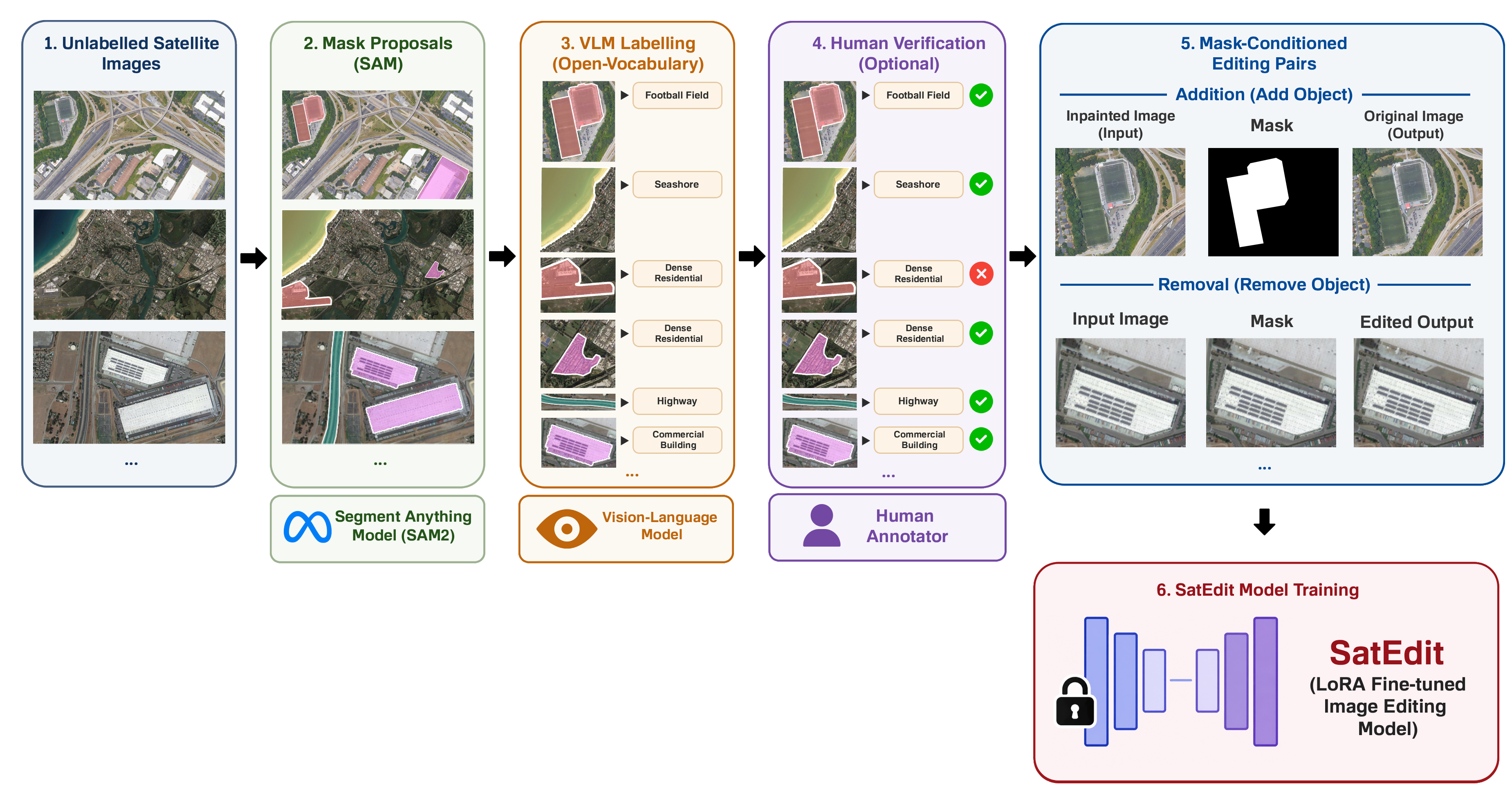}
\caption{\textbf{SatEdit data-generation and fine-tuning pipeline.} Unlabeled satellite imagery is converted into mask-conditioned editing supervision through SAM2 mask proposal generation, VLM-based segment labeling, lightweight human verification, and mask-guided construction of paired addition and removal examples. The resulting source image, binary mask, and text instruction are used to fine-tune the SatEdit image editing model.}
\label{fig:satedit_pipeline}
\end{figure*}

\subsection{Overview}
SatEdit starts from an unlabeled satellite image and a candidate object region. The training example pairs a source image with a binary mask and a text instruction, then asks the model to add or remove the named object inside that mask. Figure~\ref{fig:overview} shows the intended editing behavior: the model changes the selected region while leaving roads, shadows, field boundaries, and nearby structures intact. Although training uses addition and removal pairs, the same mask-conditioned behavior also supports replacement prompts qualitatively.

The data pipeline is shown in Figure~\ref{fig:satedit_pipeline}. For each image, SAM2 \cite{ravi2024sam2} proposes candidate masks. We query Qwen3-VL \cite{bai2025qwen3vl} on mask-aware crops to assign each retained segment a label from the satellite-object vocabulary, then use a lightweight verification pass to correct obvious label errors. The verified segments are converted into paired editing tasks: removal examples erase the object from the masked region, and addition examples ask the model to place the labeled object into a compatible masked region. These examples provide the source image, mask, and instruction used to fine-tune SatEdit.

\subsection{Segment-Level VLM-Assisted Annotation Pipeline}

We build segment-level annotations with a semi-automatic pipeline. A segmentation model proposes regions, a VLM assigns labels to mask-aware crops, and a short human verification pass corrects the remaining label errors.

\paragraph{Overview.}
Given an input image \( I \in \mathbb{R}^{H \times W \times 3} \), our goal is to produce a structured set of annotations:
\[
\mathcal{A} = \{(M_i, y_i, b_i)\}_{i=1}^{N}
\]
where \( M_i \in \{0,1\}^{H \times W} \) is a binary segmentation mask, \( y_i \in \mathcal{Y} \) is a semantic label from a predefined vocabulary, and \( b_i \in \mathbb{R}^4 \) is the corresponding bounding box.

The pipeline consists of three stages: (1) mask proposal generation, (2) segment-level semantic labeling via a vision-language model (VLM), and (3) human verification.

\paragraph{Stage 1: Mask Proposal Generation via SAM2.}
We first generate a set of candidate object segments using SAM2's automatic mask generator \cite{ravi2024sam2}:
\[
\mathcal{M} = \texttt{SAM2}(I)
\]

SAM2 produces dense, class-agnostic masks that transfer well to overhead imagery. The raw outputs include many small or redundant regions, so we apply area-based filtering:
\[
\mathcal{M}' = \left\{ M \in \mathcal{M} \;\middle|\; \frac{|M|}{HW} \ge \tau \right\}
\]
where \( \tau \) is a minimum area ratio threshold (set to 0.01 in our implementation).

To further control computational cost, we subsample masks:
\[
\tilde{\mathcal{M}} = \texttt{Sample}(\mathcal{M}', K)
\]
where \( K \) is the maximum number of segments per image.

\paragraph{Stage 2: Segment-Level VLM Labeling.}
For each retained mask \( M_i \), we extract a mask-aware crop:
\[
C_i = I[b_i] \odot M_i
\]
where \( b_i \) is the tight bounding box of \( M_i \), and pixels outside the mask are set to a neutral value.

Each cropped segment is then passed to a vision-language model \( f_{\theta} \):
\[
y_i = f_{\theta}(C_i, \mathcal{Y})
\]
where \( \mathcal{Y} \) is a predefined taxonomy of satellite-relevant classes.

\textbf{Prompt Design.} We formulate labeling as a constrained classification task by restricting outputs to the predefined vocabulary \( \mathcal{Y} \). This reduces hallucinations and enforces consistency.

Unlike prior work that applies VLMs on full images, our segment-level querying reduces multi-object ambiguity and improves localization-label alignment.

\paragraph{Stage 3: Lightweight Human Verification.}
To correct residual errors from the VLM, we introduce a single-pass human verification step:
\[
\hat{y}_i = \texttt{Verify}(y_i)
\]

Annotators are presented with the image, mask overlay, and predicted label, and can either accept or correct it. This reduces annotation effort to validation rather than creation, enabling efficient large-scale labeling.

\paragraph{Output Representation.}
Each annotated segment is stored as:
\[
(M_i, y_i, b_i, r_i)
\]
where \( r_i \) is the run-length encoding (RLE) of the mask for compact storage and compatibility with standard benchmarks.

\paragraph{Implementation Details.}
We incorporate several practical optimizations:
\begin{itemize}
    \item Mask-aware cropping with background neutralization
    \item COCO-style RLE encoding for masks
    \item Randomized mask sampling to reduce bias
    \item Visualization overlays for quality inspection
\end{itemize}

\paragraph{Discussion.}
Compared to manual annotation, our pipeline offers:
\begin{itemize}
    \item Scalability via automated segmentation and labeling
    \item Semantic richness through object-level annotations
    \item Cost efficiency by minimizing human effort
    \item Modularity for independent component improvements
\end{itemize}

\paragraph{Limitations.}
\begin{itemize}
    \item Small objects may be filtered out due to area thresholding
    \item VLM performance depends on prompt design and vocabulary coverage
    \item Mask quality directly impacts classification accuracy
    \item Quantitative comparison against closed-source API systems is limited to a modest test subset because each additional baseline query incurs monetary cost
\end{itemize}

\paragraph{Dataset Construction.}
Applying this pipeline to satellite imagery transforms image-level datasets into dense, segment-level annotated datasets suitable for detection, segmentation, and multimodal reasoning tasks.

\subsection{Dataset Construction for Mask-Conditioned Editing}
We convert the verified segments into training pairs for a mask-conditioned editor. Each pair contains a source image, a binary mask, an instruction, and a target image.

\begin{itemize}
    \item \textbf{Task generation.} For each segment, we randomly assign it to an \emph{addition} or \emph{removal} task.  
        \begin{itemize}
            \item \emph{Removal:} The original image is the source image. The target image is generated by inpainting the segment with LaMA.
            \item \emph{Addition:} The target image contains the object. The source image is the inpainted version with the object removed.
        \end{itemize}
    \item \textbf{Mask conditioning.} The segment mask is provided as an explicit model input, so the instruction is tied to a specific region.
    \item \textbf{Sampling.} Random sampling across images exposes the model to different object classes, locations, and local contexts.
\end{itemize}

The resulting examples teach the model which region to edit, what semantic change to make, and what surrounding context to preserve.

\subsection{Mask-Conditioned Model Training}
The SatEdit model is trained on the constructed dataset using Low-Rank Adaptation (LoRA) to efficiently adapt a pre-trained high-resolution image editing backbone. The complete fine-tuning configuration, including adapter rank, optimizer, scheduler, sampling settings, and model quantization options, is provided in Appendix~\ref{app:fine_tuning_details}.

\[
\mathcal{L}_{\text{total}} = \lambda_1 \mathcal{L}_{\text{recon}} + \lambda_2 \mathcal{L}_{\text{semantic}} + \lambda_3 \mathcal{L}_{\text{adv}}
\]

\begin{itemize}
    \item \textbf{Reconstruction Loss} $\mathcal{L}_{\text{recon}}$: Encourages the model output to match the target image at a pixel level.
    \item \textbf{Semantic Consistency Loss} $\mathcal{L}_{\text{semantic}}$: Ensures that the edited object aligns with the intended semantic label, measured via VLM embeddings.
    \item \textbf{Adversarial Loss} $\mathcal{L}_{\text{adv}}$: Encourages realism in edited regions.
\end{itemize}

By training with mask-conditioned pairs, SatEdit learns to place, remove, or modify objects precisely according to the mask, enabling fine-grained, generalizable editing operations across unseen object categories and editing scenarios.

\begin{figure*}[t]
\centering
\includegraphics[width=0.82\textwidth]{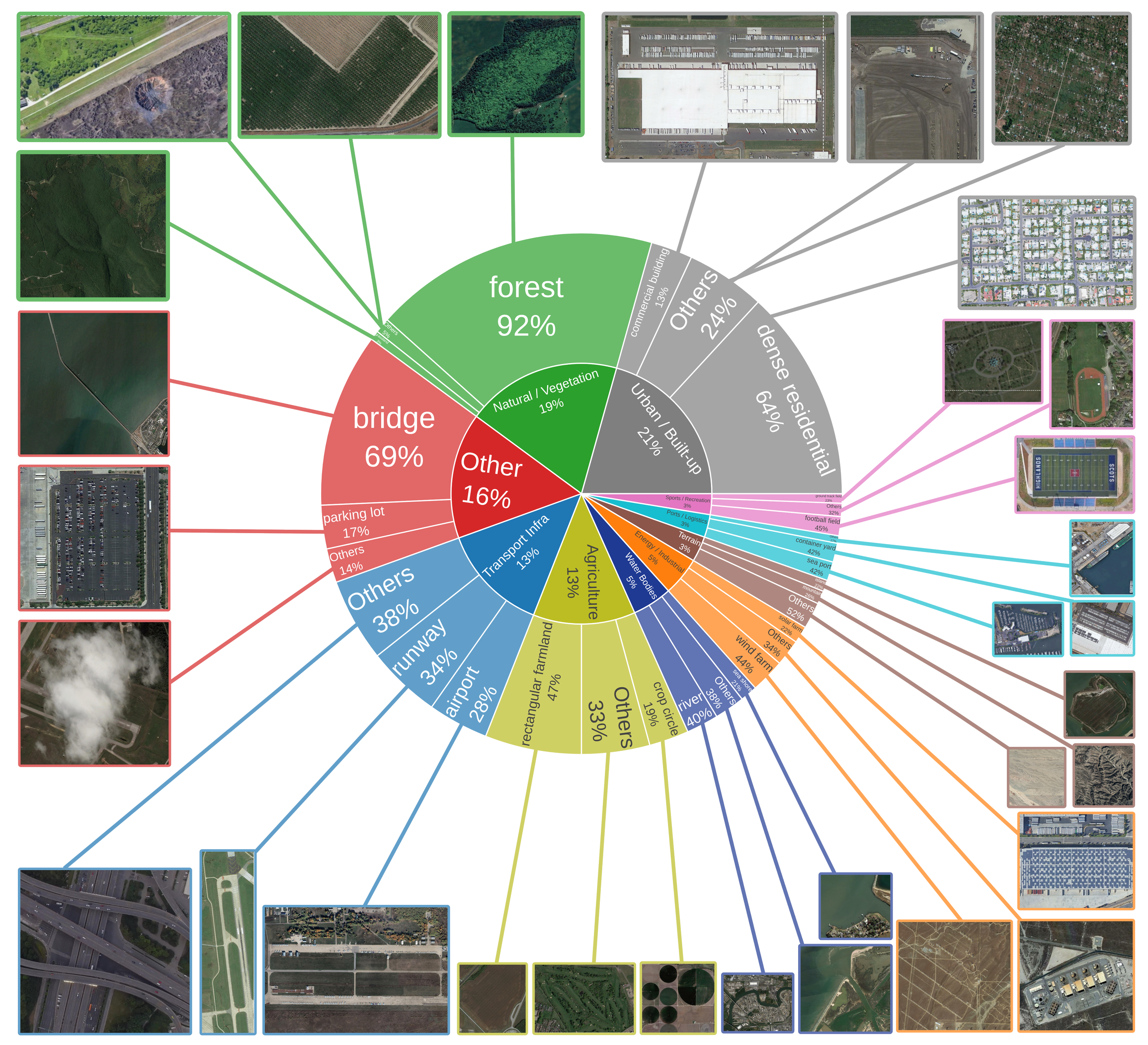}
\caption{\textbf{Semantic composition of the SODA-A-derived SatEdit annotations.} The sunburst groups the 91 segment labels into broad satellite-scene categories. For readability, each group shows its two most frequent classes explicitly, while the remaining labels are collapsed into \emph{Others}. Percentages are computed within each parent group.}
\label{fig:sunburst_class_balance}
\end{figure*}

\section{Experiments and Results}
\label{sec:experiments}

We evaluate whether SatEdit edits the requested object inside the mask while preserving the surrounding satellite scene. The experiments compare SatEdit with Nano Banana 2, GPT Image 2, and Qwen-Image-Edit on addition and removal prompts, then examine qualitative generalization and LoRA scale sensitivity.

\subsection{Datasets}
We use the SODA-A satellite imagery dataset from the Satellite Image Deep Learning collection on Hugging Face.\footnote{\url{https://huggingface.co/datasets/satellite-image-deep-learning/SODA-A/tree/main}} SODA-A contains high-resolution aerial scenes from several land-use categories, which gives the editing benchmark a mix of urban, agricultural, transportation, and natural contexts.

The released SODA-A images do not include the mask-conditioned editing pairs needed for this work. We generate those annotations with the SatEdit pipeline: SAM proposes object masks, the VLM assigns semantic labels, and verified segments are converted into paired \emph{addition} and \emph{removal} examples. Table~\ref{tab:dataset_stats} summarizes the resulting dataset: 1,014 images, 586 null examples, and 852 object annotations, averaging 0.8 annotations per image across 91 classes.

\begin{table}[t]
\centering
\caption{\textbf{SODA-A-derived SatEdit dataset statistics.} Annotations are generated by the SatEdit pipeline from SODA-A imagery.}
\label{tab:dataset_stats}
\begin{tabular}{lc}
\toprule
\textbf{Statistic} & \textbf{Value} \\
\midrule
Images & 1,014 \\
Null examples & 586 \\
Object annotations & 852 \\
Annotations per image & 0.8 \\
Classes & 91 \\
\bottomrule
\end{tabular}
\end{table}

Figure~\ref{fig:sunburst_class_balance} shows the label distribution at two levels. The inner ring groups labels into broad satellite-scene categories, and the outer ring shows the most frequent labels within each group. The distribution is long-tailed: common labels such as forest, dense residential areas, bridges, rectangular farmland, runways, and airports provide repeated supervision, while the grouped \emph{Others} slices keep rarer classes in the dataset. This matters because a mask-conditioned editor should learn both common spatial patterns and less frequent object appearances.

\vspace*{2\baselineskip}
\subsection{Baselines}
We compare against three image editing systems:
\begin{itemize}
    \item \textbf{GPT Image 2:} a proprietary GPT-based image editor evaluated with the same prompts and masks.
    \item \textbf{Nano Banana 2 (Gemini 3.1 Flash Preview):} a proprietary image editing system.
    \item \textbf{Qwen-Image-Edit-2511:} an open-source VLM-driven image editing model.
\end{itemize}

\subsection{Evaluation Metrics}
We use metrics that separate prompt following, edit locality, and background preservation:

\textbf{1. Semantic Alignment:}
\begin{itemize}
    \item \textbf{CLIP Score} – similarity between the edited masked region and the text prompt.
    \item \textbf{CLIP Delta} – improvement in CLIP score from the original image to the edited image.
\end{itemize}

\textbf{2. Edit Localization:}
\begin{itemize}
    \item \textbf{Edit Strength} – mean absolute RGB difference inside the target mask.
    \item \textbf{Leakage} – ratio of change outside the target mask to change inside the target mask; lower values indicate better localization.
\end{itemize}

\textbf{3. Image Preservation:}
\begin{itemize}
    \item \textbf{Full-image SSIM} – structural similarity over the full luminance image.
    \item \textbf{Outside-mask PSNR} – PSNR computed only over pixels outside the edited mask.
\end{itemize}

For CLIP evaluation, we use OpenCLIP ViT-B/32 with OpenAI weights. To isolate the target object or region, pixels outside the mask are replaced by the image mean color before computing image--text similarity. We report the weighted CLIP score with weight 2.5 and clamp negative similarities to zero. All edited outputs are resized to the original image resolution before evaluation. SSIM and PSNR are computed on the luminance channel; SSIM is evaluated over the full image, while PSNR is evaluated only outside the edit mask to measure background preservation.

\begin{table*}[!b]
\centering
\footnotesize
\setlength{\tabcolsep}{3pt}
\renewcommand{\arraystretch}{0.95}

% -------------------------
% Addition
% -------------------------
\begin{tabular}{lrrrrrr}
    \toprule
    \multicolumn{7}{c}{\textbf{Object Addition}} \\
    \midrule
    Method & CLIP$\uparrow$ & CLIP $\Delta \uparrow$ & Edit Strength$\downarrow$ & Leakage$\downarrow$ & SSIM$\uparrow$ & PSNR-out$\uparrow$ \\
    \midrule
    GPT Image 2 & 0.5783 & 0.0234 & 0.0677 & 8.4240 & 0.2440 & 14.3220 \\
    Nano Banana 2 & 0.6296 & 0.0755 & \textbf{0.0464} & \textbf{6.5797} & \textbf{0.2556} & \textbf{14.3961} \\
    Qwen-Image-Edit-2511 & 0.6306 & 0.0765 & 0.0515 & 10.2330 & 0.2275 & 13.5533 \\
    \textbf{SatEdit (ours)} & \textbf{0.6371} & \textbf{0.0829} & 0.0475 & 7.4302 & 0.2345 & 13.7552 \\
\midrule
\end{tabular}

\vspace{0.25em}

% -------------------------
% Removal
% -------------------------
\begin{tabular}{lrrrrrr}
\toprule
\multicolumn{7}{c}{\textbf{Object Removal}} \\
\midrule
Method & CLIP$\uparrow$ & CLIP $\Delta \uparrow$ & Edit Strength$\downarrow$ & Leakage$\downarrow$ & SSIM$\uparrow$ & PSNR-out$\uparrow$ \\
\midrule
GPT Image 2 & 0.5971 & -0.0296 & 0.0231 & 5.2809 & 0.3048 & 14.3068 \\
Nano Banana 2 & 0.5892 & -0.0006 & 0.0228 & \textbf{2.2194} & \textbf{0.6382} & \textbf{22.1090} \\
Qwen-Image-Edit-2511 & 0.6068 & 0.0170 & \textbf{0.0176} & 5.2550 & 0.4211 & 16.6512 \\
\textbf{SatEdit (ours)} & \textbf{0.6125} & \textbf{0.0315} & 0.0349 & 5.2349 & 0.3377 & 14.0632 \\
\midrule
\end{tabular}

\vspace{0.25em}

% -------------------------
% Aggregate
% -------------------------
\begin{tabular}{lrrrrrr}
\toprule
\multicolumn{7}{c}{\textbf{Aggregate Results}} \\
\midrule
Method & CLIP$\uparrow$ & CLIP $\Delta \uparrow$ & Edit Strength$\downarrow$ & Leakage$\downarrow$ & SSIM$\uparrow$ & PSNR-out$\uparrow$ \\
\midrule
GPT Image 2 & 0.5821 & 0.0128 & 0.0588 & 7.7954 & 0.2561 & 14.3190 \\
Nano Banana 2 & 0.6233 & 0.0635 & \textbf{0.0427} & \textbf{5.8913} & \textbf{0.3160} & \textbf{15.6139} \\
Qwen-Image-Edit-2511 & 0.6268 & 0.0671 & 0.0461 & 9.4470 & 0.2580 & 14.0424 \\
\textbf{SatEdit (ours)} & \textbf{0.6322} & \textbf{0.0726} & 0.0450 & 6.9911 & 0.2552 & 13.8168 \\
\bottomrule
\end{tabular}

\caption{\textbf{Quantitative comparison of mask-conditioned satellite image editing methods.} SatEdit has the highest CLIP and CLIP $\Delta$ scores across addition, removal, and aggregate settings. Nano Banana 2 changes the background least, with the best preservation scores and lowest leakage.}
\label{tab:combined_results}
\end{table*}

\subsection{Quantitative Benchmark Results}
The quantitative benchmark uses 20 images from the test split. We evaluate object \emph{addition}, object \emph{removal}, and the aggregate over both tasks. For addition, target labels are sampled from the full class vocabulary rather than only from labels already present in the source image. Several held-out labels are reserved for qualitative stress tests. Because two baselines are closed-source API systems, the benchmark is deliberately small; we use it to compare relative behavior under controlled prompts and masks, not to estimate deployment-level performance.

\paragraph{Addition.}
For object insertion, SatEdit has the highest CLIP score and CLIP delta in Table~\ref{tab:combined_results}. This indicates stronger alignment between the edited mask region and the requested class. Nano Banana 2 and Qwen-Image-Edit are close on CLIP, but Qwen-Image-Edit changes more outside the mask. GPT Image 2 preserves the background competitively, though its prompt-directed improvement is weaker than SatEdit's.

\paragraph{Removal.}
For removal, SatEdit again has the highest CLIP score and CLIP delta. Nano Banana 2 is more conservative. It has the lowest leakage and the best SSIM and PSNR-out, but it moves the masked region less semantically than SatEdit. Conservative removals can score well on reconstruction metrics even when the requested object change is incomplete, so this tradeoff matters when reading the table.

\paragraph{Aggregate.}
Across both tasks, SatEdit has the highest aggregate CLIP score and CLIP delta. Nano Banana 2 changes the surrounding image least and has the strongest preservation scores. The aggregate result supports a narrower claim: SatEdit improves prompt-aligned object editing inside the mask, while Nano Banana 2 remains the more conservative editor.

\begin{figure*}[!t]
\centering
\includegraphics[width=0.96\textwidth]{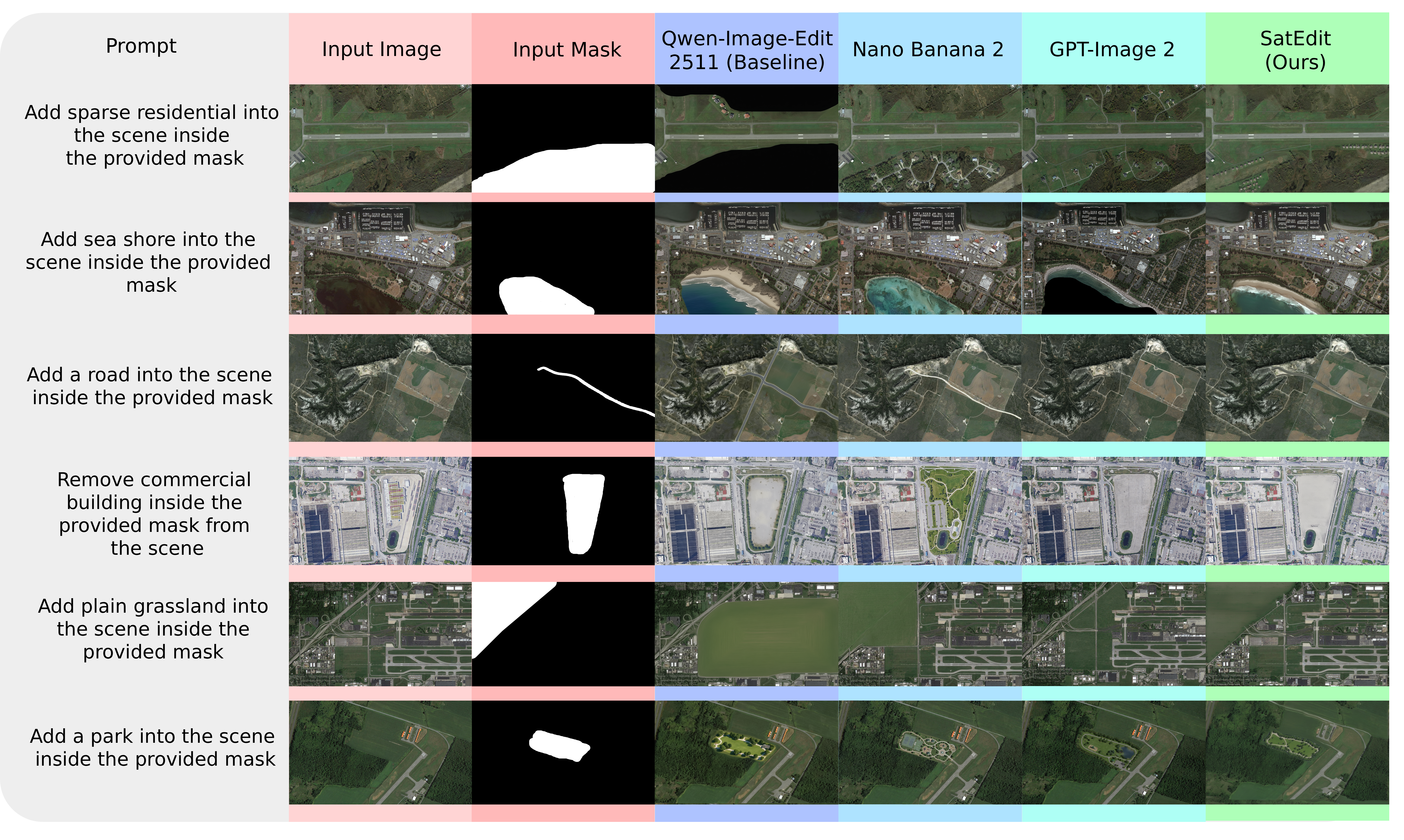}
\caption{Qualitative comparison of editing results. Columns show the editing prompt, input image, input mask, baseline outputs from Qwen-Image-Edit-2511, Nano Banana 2, and GPT Image 2, and the SatEdit output.}
\label{fig:examples}
\end{figure*}

\subsection{Qualitative Results}
Figure~\ref{fig:examples} compares representative edits. Across the examples, SatEdit preserves the input mask and confines the requested semantic change to the masked region. The baselines more often change pixels outside the mask, introduce objects at implausible scale or viewpoint, or leave the requested semantic change incomplete.

The first row shows a sparse residential insertion. Qwen-Image-Edit-2511 does not respect the mask and turns the selected region nearly black. Nano Banana 2 and SatEdit both place residential structure inside the masked area, but GPT Image 2 also edits the unmasked upper half of the image by adding extra houses outside the requested region. In the shoreline example, Qwen-Image-Edit-2511 and GPT Image 2 again spill beyond the mask boundary, whereas SatEdit and Nano Banana 2 keep the edit localized. For the road-addition prompt, Qwen-Image-Edit-2511 adds an additional road perpendicular to the target region, Nano Banana 2 produces a white strip that does not resemble a road in the surrounding satellite context, and GPT Image 2 leaves the image largely unchanged. SatEdit adds a road only within the masked region.

The removal and replacement examples show a similar pattern. In the commercial-building removal case, all methods remove the target building, but Nano Banana 2 introduces an unsolicited park-like region, while the other methods replace the removed structure with a more neutral concrete-like plot. For the plain-grassland prompt, Qwen-Image-Edit-2511 modifies areas outside the mask, Nano Banana 2 also spills beyond the target region, and GPT Image 2 makes little visible change; SatEdit keeps the grassland edit inside the mask. The final row evaluates a class not used during SatEdit training. All methods insert a park-like region, but Qwen-Image-Edit-2511 produces an output that appears visually inconsistent with the top-down satellite viewpoint, with an oblique-looking structure. SatEdit preserves the overhead perspective and keeps the edit localized, suggesting that mask-conditioned supervision improves spatial control even for unseen semantic classes.

\subsection{Qualitative Generalization to Unseen Classes}
For unseen classes, we add held-out labels to the prompt pool and inspect the outputs. We do not report a separate quantitative table for these cases because the sample is small and includes costly closed-source baselines. Instead, Figure~\ref{fig:examples} treats them as stress tests: does the method respect the mask, preserve the surrounding scene, and produce a plausible object or removal?

\subsection{LoRA Strength Ablation}
The current SatEdit model is trained on the final human-verified annotations from the full pipeline. We do not ablate away the annotation pipeline itself, since training on unverified VLM labels or removing SAM masks would test a different data source rather than the final model. Instead, we vary the LoRA scale, which controls how much the fine-tuned satellite-editing adapter contributes at inference time.

At inference time, we vary the LoRA scale \(\alpha\) while keeping the base model, prompt, mask, and random seed fixed. The edited image is generated as:
\[
\theta_{\alpha} = \theta_{\text{base}} + \alpha \Delta\theta_{\text{LoRA}},
\]
where \(\alpha=0\) corresponds to the frozen base model and larger values increase the contribution of the SatEdit adapter. This ablation isolates whether the adapter improves satellite-specific mask alignment or merely increases the overall edit magnitude.

\begin{table*}[!b]
\centering
\scriptsize
\setlength{\tabcolsep}{3pt}
\renewcommand{\arraystretch}{0.95}
\begin{tabular}{>{\columncolor{blue!7}}r>{\columncolor{blue!7}}r>{\columncolor{blue!7}}r>{\columncolor{blue!7}}r>{\columncolor{blue!7}}r>{\columncolor{blue!7}}r>{\columncolor{blue!7}}r@{\hspace{1.6em}}>{\columncolor{orange!10}}r>{\columncolor{orange!10}}r>{\columncolor{orange!10}}r>{\columncolor{orange!10}}r>{\columncolor{orange!10}}r>{\columncolor{orange!10}}r>{\columncolor{orange!10}}r}
\toprule
\multicolumn{7}{c}{\cellcolor{blue!13}\textbf{Lower adapter scales}} &
\multicolumn{7}{c}{\cellcolor{orange!16}\textbf{Higher adapter scales}} \\
\midrule
\(\alpha\) & CLIP\(\uparrow\) & CLIP \(\Delta \uparrow\) & Strength\(\downarrow\) & Leak.\(\downarrow\) & SSIM\(\uparrow\) & PSNR-out\(\uparrow\) &
\(\alpha\) & CLIP\(\uparrow\) & CLIP \(\Delta \uparrow\) & Strength\(\downarrow\) & Leak.\(\downarrow\) & SSIM\(\uparrow\) & PSNR-out\(\uparrow\) \\
\midrule
0.0 & 0.6220 & 0.0590 & 0.0223 & 10.8788 & 0.3899 & 17.3227 &
1.1 & 0.6051 & 0.0421 & 0.0178 & 8.7313 & 0.3907 & 18.2481 \\
0.1 & \textbf{0.6272} & \textbf{0.0642} & 0.0213 & 10.6218 & 0.3932 & 17.6036 &
1.2 & 0.5908 & 0.0278 & 0.0178 & 8.8191 & 0.3915 & 18.1444 \\
0.2 & 0.6160 & 0.0530 & 0.0195 & 10.6124 & 0.3967 & 17.9694 &
1.3 & 0.5765 & 0.0135 & 0.0178 & 9.6599 & 0.3863 & 17.5862 \\
0.3 & 0.6044 & 0.0414 & 0.0187 & 11.1657 & 0.4013 & 18.8208 &
1.4 & 0.5787 & 0.0157 & \textbf{0.0166} & 10.5098 & 0.3827 & 17.3266 \\
0.4 & 0.6115 & 0.0485 & 0.0191 & 11.1718 & \textbf{0.4070} & \textbf{18.9936} &
1.5 & 0.5755 & 0.0125 & 0.0174 & 10.9491 & 0.3799 & 17.1244 \\
0.5 & 0.6209 & 0.0579 & 0.0171 & 11.3480 & 0.4037 & 18.9076 &
1.6 & 0.5705 & 0.0075 & 0.0213 & 10.3809 & 0.3688 & 16.5689 \\
0.6 & 0.6259 & 0.0629 & 0.0172 & 7.7512 & 0.3982 & 18.7091 &
1.7 & 0.5627 & -0.0004 & 0.0227 & 10.4278 & 0.3676 & 16.1945 \\
0.7 & 0.6229 & 0.0599 & 0.0184 & \textbf{7.3775} & 0.3986 & 18.5470 &
1.8 & 0.5593 & -0.0037 & 0.0234 & 10.4569 & 0.3661 & 15.9290 \\
0.8 & 0.6113 & 0.0483 & 0.0188 & 7.6910 & 0.3956 & 18.5489 &
1.9 & 0.5622 & -0.0008 & 0.0236 & 10.1758 & 0.3632 & 15.7816 \\
0.9 & 0.6093 & 0.0463 & 0.0187 & 8.3171 & 0.3927 & 18.4058 &
2.0 & 0.5644 & 0.0014 & 0.0237 & 9.9661 & 0.3599 & 15.6408 \\
1.0 & 0.6116 & 0.0486 & 0.0179 & 8.5576 & 0.3954 & 18.4233 &
\multicolumn{7}{c}{} \\
\bottomrule
\end{tabular}
\caption{\textbf{LoRA-strength ablation.} The same prompt, mask, seed, and base model are used while varying only the SatEdit adapter scale \(\alpha\). Low-to-moderate scales give the best CLIP alignment, \(\alpha=0.7\) minimizes leakage, and high scales reduce both CLIP alignment and preservation scores.}
\label{tab:lora_strength_ablation}
\end{table*}

Table~\ref{tab:lora_strength_ablation} sweeps \(\alpha \in [0,2]\) in increments of 0.1. The effect is non-monotonic. Low-to-moderate scales give the best semantic alignment: \(\alpha=0.1\) has the highest CLIP score and CLIP \(\Delta\), while \(\alpha=0.6\) remains close and substantially reduces leakage. Background preservation peaks around \(\alpha=0.3\)--\(0.5\), with the best SSIM and outside-mask PSNR at \(\alpha=0.4\). Once \(\alpha\) exceeds 1.2, semantic alignment weakens; CLIP \(\Delta\) approaches zero and becomes negative at \(\alpha=1.7\)--\(1.9\). In this range, the adapter appears to overpower useful structure from the base model. Edit strength also rises while SSIM and PSNR fall, which matches the visual pattern of over-adaptation.

The most balanced range is \(\alpha \approx 0.6\)--\(0.7\). This range gives the lowest leakage in the sweep while keeping CLIP alignment and background fidelity competitive. If semantic alignment alone is the priority, \(\alpha=0.1\) is best; if preservation alone is the priority, \(\alpha=0.4\) is best.

\begin{figure*}[!t]
\centering
\includegraphics[width=0.92\textwidth]{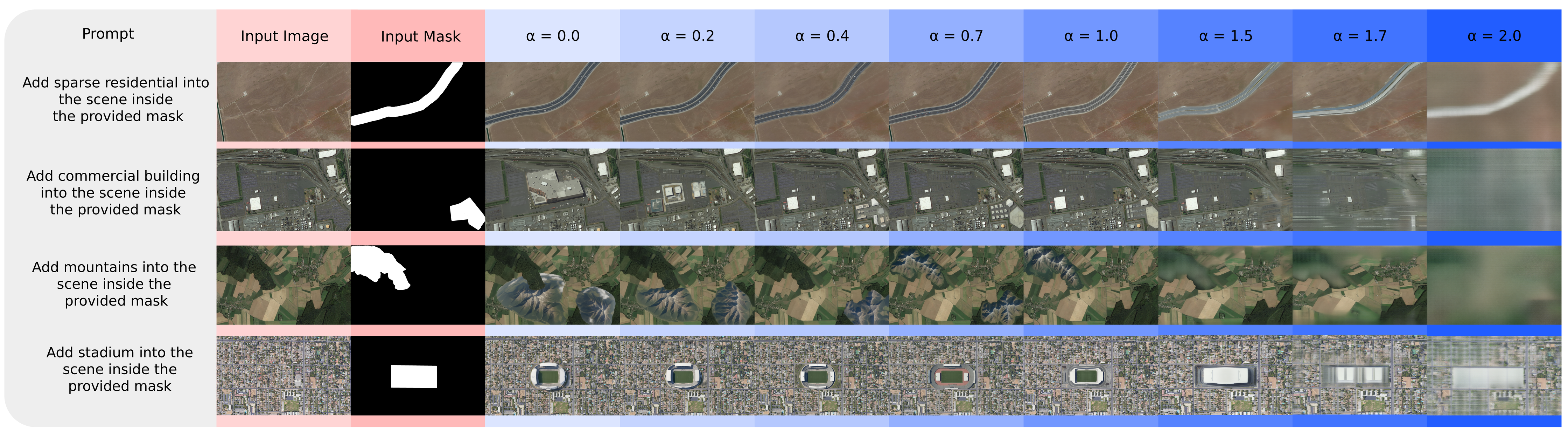}
\caption{\textbf{Qualitative effect of LoRA scale.} Each row fixes the prompt, input image, mask, and seed while varying only the SatEdit adapter scale \(\alpha\). Low-to-moderate scales preserve the satellite context and produce localized edits, while high scales increasingly blur the scene, wash out object boundaries, or replace the masked target with broad texture-like changes.}
\label{fig:lora_alpha_qualitative}
\end{figure*}

The qualitative grid in Figure~\ref{fig:lora_alpha_qualitative} explains the non-monotonic trend. At \(\alpha=0.0\)--\(0.4\), the base prior dominates. Edits are conservative and often preserve roads, fields, and urban texture, but the requested object can be weak or incomplete. Around \(\alpha=0.7\)--\(1.0\), the adapter contributes enough satellite-specific structure to make object insertions visible while keeping most changes inside the mask. At larger scales, especially \(\alpha=1.5\)--\(2.0\), the adapter begins to overpower the base model. Outputs become smoother and less geographically faithful, with blurred roads, washed-out buildings, and masked regions that resemble generic texture rather than the requested object. This visual pattern matches the quantitative result: moderate adapter strength gives the best tradeoff between semantic edit strength, leakage, and background preservation.

\FloatBarrier

\section{Conclusion}
\label{sec:conclusion}
SatEdit tackles the data problem behind mask-conditioned satellite image editing. Rather than manually authoring before/after pairs, it builds supervision from unlabeled overhead imagery: SAM2 proposes candidate masks \cite{ravi2024sam2}, a VLM assigns segment labels, annotators verify the labels, and inpainting turns the verified segments into addition and removal pairs. This keeps human effort on label checking while retaining the object masks needed for localized edits.

On the SODA-A-derived benchmark, this supervision improves prompt alignment inside the target region. SatEdit obtains the best aggregate CLIP score (0.6322) and CLIP delta (0.0726) among the evaluated systems, while the preservation metrics show the expected tradeoff between stronger edits and background fidelity. Qualitative examples show the same pattern: SatEdit usually confines the requested change to the mask and preserves the surrounding satellite context better than methods that ignore or spill beyond the selected region.

The method is still limited by the size of the generated dataset, the cost of evaluating closed-source baselines, and errors inherited from segmentation and inpainting. These are practical constraints rather than settled properties of the approach. The main result is that verified segment-level labels are enough to train a spatially controlled satellite editor, which makes VLM-assisted annotation a useful route for building larger remote-sensing editing datasets.

\section{Future Work}
Future versions of SatEdit should start with more varied satellite imagery. The current dataset covers several scene types, but it is still too small to capture the full range of sensors, resolutions, geographies, seasons, and rare object classes that appear in operational remote-sensing data. A larger dataset would also reduce the current class imbalance and give the editor more examples of small objects, dense urban layouts, agricultural structure, and disaster-related changes.

The data-generation pipeline also needs stricter quality control. Better filtering of SAM2 proposals, confidence-aware VLM labeling, and more systematic human verification would reduce noisy training pairs before fine-tuning. This matters because label mistakes and poor masks affect both sides of the task: they can teach the model the wrong object class, and they can weaken the link between the instruction and the intended edit region.

SatEdit should also move beyond single-object addition and removal. Useful satellite editing often requires more specific instructions, such as changing object attributes, editing several regions at once, or answering follow-up instructions in a conversational workflow. The same framework could also be adapted to multispectral and hyperspectral imagery, where an edit should remain plausible not only in RGB appearance but also in spectral response.

Finally, evaluation should expand beyond the small controlled benchmark used here. Larger test sets and task-specific studies for urban planning, land-use monitoring, environmental assessment, and disaster response would make it easier to judge when the editor is useful. Future metrics should measure geographic plausibility, boundary consistency, object realism, and preservation of nearby infrastructure, not only prompt alignment and pixel-level preservation.

\FloatBarrier
\clearpage

\appendix
\section{SatEdit Fine-Tuning Details}
\label{app:fine_tuning_details}

\paragraph{Training toolkit.}
We fine-tuned SatEdit with Ostris AI Toolkit (\url{https://github.com/ostris/ai-toolkit}), an open-source training suite for diffusion models that supports Qwen-Image-Edit-2511. The settings below summarize the run used for the experiments in this paper.

\paragraph{Compute.}
Training ran on a rented NVIDIA RTX A6000 GPU with CUDA. We used the toolkit's low-memory options, including gradient checkpointing, text-embedding caching, latent caching to disk, and quantized model loading.

\paragraph{Base model.}
We fine-tuned Qwen-Image-Edit-2511 with the \texttt{qwen\_image\_edit\_plus:2511} architecture setting. The image-editing backbone and text encoder were loaded with \texttt{qfloat8} quantization. We also enabled the toolkit option that matches the model to the target training resolution.

\paragraph{Adapter configuration.}
We trained a LoRA adapter instead of updating the full model. The adapter rank was 64 for both linear and convolutional layers, and LoRA alpha was also 64 for both layer types. We did not exclude modules through name-based filtering. The final adapter was saved in \texttt{diffusers} format with \texttt{bf16} precision.

\paragraph{Dataset inputs.}
Training examples were loaded from the SatEdit target-image folder, with paired controls provided through two control-image directories. The first control image was the source satellite image, and the second was the spatial mask/control image. The training resolution was \(1024\), and each example used one frame. Captions were read from \texttt{.txt} files with caption dropout set to 0.05. Latents were cached to disk to avoid repeated encoder computation.

\paragraph{Optimization.}
We trained for 12,000 optimization steps with batch size 1 and gradient accumulation 2, giving an effective batch size of 2. The optimizer was 8-bit AdamW with learning rate \(5 \times 10^{-5}\) and weight decay 0. The noise scheduler was \texttt{flowmatch}, timesteps used weighted sampling, and the objective was mean squared error. We trained the image-editing network and kept the text encoder frozen. Training used \texttt{bf16} precision and no EMA.

\paragraph{Guidance and sampling.}
Differential guidance was enabled during training with scale 1.2. We generated validation samples every 250 steps and saved checkpoints at the same interval, retaining the four most recent step checkpoints. Validation used the \texttt{flowmatch} sampler at \(1280 \times 720\) resolution with 40 sampling steps, guidance scale 4, seed 42, and seed walking enabled. The validation prompts covered representative insertion and removal requests for commercial buildings, dense residential areas, highways, dense forests, and seaports with boats.

\end{document}